\documentclass[letterpaper, 10 pt, conference]{ieeeconf}  % Comment this line out if you need a4paper

\IEEEoverridecommandlockouts                              % This command is only needed if 
\usepackage{graphics} % for pdf, bitmapped graphics files
\usepackage{epsfig} % for postscript graphics files
\usepackage{mathptmx} % assumes new font selection scheme installed
\usepackage{times} % assumes new font selection scheme installed
\usepackage{amsmath} % assumes amsmath package installed
\usepackage{amssymb}  % assumes amsmath package installed

\usepackage{caption}
\usepackage{subcaption}
\usepackage{booktabs}
\usepackage{xcolor}
\usepackage{soul}
\usepackage{gensymb}

\title{\LARGE \bf
Towards the Classification of Error-Related Potentials using Riemannian Geometry
}

\author{Yichen Tang$^{1,3}$, {\it Student Member, IEEE}, Jerry J. Zhang$^{1}$, {\it Student Member, IEEE}, \\Paul M. Corballis$^{2}$, and Luke E. Hallum$^{1}$, {\it Member, IEEE}% <-this % stops a space

\thanks{*This work was partly supported by a University of Auckland Faculty of Engineering Research Development Fund award to L.E.H.}% <-this % stops a space
\thanks{$^{1}$Yichen Tang, Jerry J. Zhang, and Luke E. Hallum are with Department of Mechanical Engineering, University of Auckland, Auckland, New Zealand}%
\thanks{$^{2}$Paul M. Corballis is with School of Psychology, University of Auckland, Auckland, New Zealand}%
\thanks{$^{3}$Yichen Tang is the corresponding author. Email: {\tt\small ytan415@aucklanduni.ac.nz}}
}

\begin{document}

\maketitle
\thispagestyle{empty}
\pagestyle{empty}

%%%%%%%%%%%%%%%%%%%%%%%%%%%%%%%%%%%%%%%%%%%%%%%%%%%%%%%%%%%%%%%%%%%%
\begin{abstract}
The error-related potential (ErrP) is an event-related potential (ERP) evoked by an experimental participant's recognition of an error during task performance. ErrPs, originally described by cognitive psychologists, have been adopted for use in brain-computer interfaces (BCIs) for the detection and correction of errors, and the online refinement of decoding algorithms. Riemannian geometry-based feature extraction and classification is a new approach to BCI which shows good performance in a range of experimental paradigms, but has yet to be applied to the classification of ErrPs. Here, we describe an experiment that elicited ErrPs in seven normal participants performing a visual discrimination task. Audio feedback was provided on each trial. We used multi-channel electroencephalogram (EEG) recordings to classify ErrPs (success/failure), comparing a Riemannian geometry-based method to a traditional approach that computes time-point features. Overall, the Riemannian approach outperformed the traditional approach (78.2\% versus 75.9\% accuracy, p \textless 0.05); this difference was statistically significant (p \textless 0.05) in three of seven participants. These results indicate that the Riemannian approach better captured the features from feedback-elicited ErrPs, and may have application in BCI for error detection and correction.
\end{abstract}

%%%%%%%%%%%%%%%%%%%%%%%%%%%%%%%%%%%%%%%%%%%%%%%%%%%%%%%%%%%%%%%%%%%%%%%%%%%%%%%%
\section{INTRODUCTION}

A key goal of brain-computer interface (BCI) research is to use signals derived from the electroencephalogram (EEG) to interface with a device \cite{lotte2018review}. Typical BCIs use machine learning to translate EEG activity into control signals \cite{lotte2018review}. Because both humans and machines make mistakes \cite{chavarriaga2014errare,zeyl2016adaptive, schalk2000eeg}, error detection and correction are vital for improving the utility of a BCI \cite{chavarriaga2014errare, parra2003response}. The error-related potential (ErrP) is family of event-related potential (ERP) components that can be derived from EEG activity. ErrPs are usually evoked when a person recognises an error, regardless of whether that error was made by the person, someone else, or the BCI \cite{chavarriaga2014errare,zeyl2016adaptive,schalk2000eeg,miltner1997event}. ErrPs were originally described in the context of cognitive psychology \cite{zeyl2016adaptive,miltner1997event}, and were thereafter adopted for  BCIs for the error detection and correction and for online refinement of decoding algorithms \cite{chavarriaga2014errare,llera2011use}. ErrPs comprise a characteristic series of voltage deflections, including a frontocentral negativity followed by a positivity, and then a parietal positivity \cite{chavarriaga2014errare,ullsperger2014neural}. However, based on the source of error information, these components may have different manifestations and nomenclature \cite{chavarriaga2014errare,ullsperger2014neural}. When the person recognises their own error, the frontal components are termed error-related negativity (ERN) and the positive deflection (Pe), while the ErrP generated by feedback is termed the feedback-related negativity (FRN).  The later frontal (P3A \cite{chavarriaga2014errare,miltner1997event,gehring1993neural}) and parietal (P3B \cite{ullsperger2014neural,polich2007updating}) positivities are cognitive components not specific to error monitoring or feedback.

Traditionally, ERP classification for BCI involves computation of features at different points in time on a number of EEG channels \cite{lotte2018review}, which are then concatenated to form a high-dimensional feature vector. By contrast, a new approach that uses Riemannian geometry \cite{lotte2018review} has often outperformed traditional methods \cite{lotte2018review,yger2016riemannian}. Instead of extracting a predefined feature vector, these methods map ERPs onto a Riemannian manifold -- often by computing a covariance matrix on individual trials -- and then classifying samples directly on the manifold, or transforming samples into vectors for classification  \cite{lotte2018review,yger2016riemannian}. Although this classification was originally designed for capturing spatial features for use in a motor-imagery paradigm, these methods can potentially be adapted to a wide range of BCI paradigms \cite{yger2016riemannian,barachant2010riemannian} by changing how the covariance matrices are computed (such as the adoption of prototype ERP responses, discussed below).

To our knowledge, Riemannian geometry-based methods have not yet been used for feature extraction and classification of ErrPs. Therefore, here we apply such a method combined with logistic regression to extract and classify ErrPs during a two-alternative forced-choice (2AFC) visual-disctimination task. We compare outcomes to a traditional feature extraction and classification method.

\section{Methods}

\subsection{Experiment and EEG Recording}

\subsubsection{Participants} \label{sec:participants}

We recorded behavioural responses and multi-channel EEG from seven participants (six males; age range: 20 to 25 years old), all with normal or corrected-to-normal vision. Participants provided informed consent. Experimental protocols were approved by the University of Auckland Human Participants Ethics Committee. 

\subsubsection{Procedure}

\begin{figure}[thpb]
      \centering
      \centerline{\includegraphics[width=0.45\textwidth]{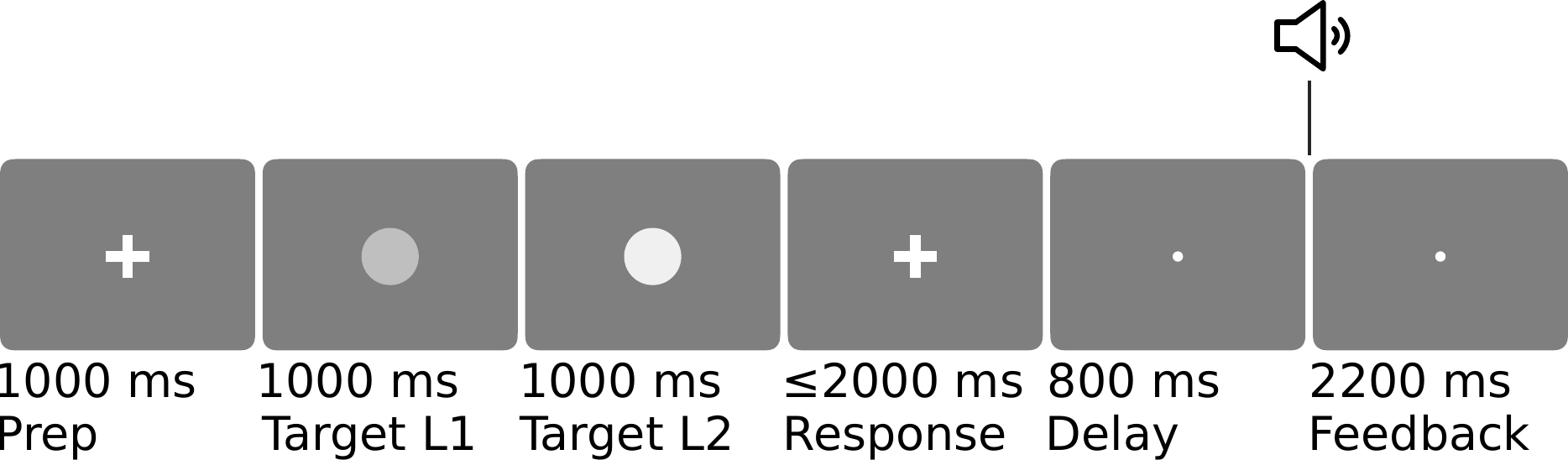}}
      \caption{Trial structure. On each trial, the participant prepared (``Prep.”) by fixating a crosshairs (width = 1\textdegree of visual angle). Then, a circular target (diameter = 0.5\textdegree) appeared in each of two 1-s temporal intervals, followed by the reappearance of the crosshairs. The participant responded with a keypress to indicate whether the 1st or 2nd target was higher luminance. Audio feedback was provided 800 ms after the participant’s response.
}
        \label{fig:trial_illustration}

   \end{figure}

Each participant sat for five blocks of 60 trials (300 trials total); blocks were separated by short breaks. Participants performed a visual discrimination task at fixation. Each trial began with a 1-s presentation of a white crosshairs on a grey background, followed by two temporal intervals. The first interval contained a circular target (diameter = 0.5\textdegree of visual angle) with luminance = L1; the second interval contained a circular target with luminance = L2. After the two intervals, the participant responded with a keypress to indicate whether the 1st or 2nd interval contained the target at higher luminance (we randomised this experimental parameter across trials). The trial structure is detailed in Fig. \ref{fig:trial_illustration}. We used a one-up/one-down staircase to adjust target luminance on a trial-by-trial basis. This staircase procedure ensured that the task was sufficiently challenging to elicit approximately equal numbers of correct and incorrect responses. On each trial, participants were allowed 2 s to provide a response before time-out; the few trials on which participants gave no response were treated as incorrect. Each block began with five practice trials which we later discarded from our analysis. We instructed participants to always fixate the display centre and to blink as infrequently as possible during the experiment. 

\subsubsection{Visual stimuli and feedback} \label{sec:stimuli}

We presented visual stimuli using two 24-inch, gamma-corrected liquid-crystal displays (60 Hz; ColorEdge CG247X; EIZO Corp., Hakusan, Japan) \cite{hallum2021liquid}, each reflected into its eye through a 45° mirror stereoscope. Stimuli at fixation were identical on both displays. (For separate experimental purposes which we describe in a companion paper \cite{zhang2021interocular}, dichoptic stimuli were presented in annuli surrounding fixation. These dichoptic stimuli were absent from participants’ awareness, and are of no consequence to the present results.) We provided correct (incorrect) feedback using 150-ms pure tone at 700 Hz (200 Hz). We used two loudspeakers (Edifier r1700bt; Edifier International Ltd., Beijing, China), each 0.55 m from its ear and 30° off the sagittal plane. The sound level of tones was adjusted to approximately 70 dB SPL at the ear. We used a phototransistor (TEPT4400; Vishay Intertechnology, Inc., Malvern, PA, United States) and a sound sensor (XC-4438; Jaycar, Rydalmere, NSW, Australia) together with two Arduino UNO R3 boards to synchronise visual and audio signals with EEG recordings.

\subsubsection{EEG recording and preprocessing}

We used a BioSemi ActiveTwo AD-box (ADC-17; ActiveTwo; Biosemi, Amsterdam, Netherlands) to record and amplify multi-channel EEG at a rate of 2048 samples per second for each of 32 scalp channels. We positioned the scalp electrodes according to the international 10-20 system; the channels were Fp1, AF3, F7, F3, FC1, FC5, T7, C3, CP1, CP5, PO7, P3, POz, PO3, O1, Oz, O2, PO4, P4, PO8, CP6, CP2, C4, T8, FC6, FC2, F4, F8, AF4, Fp2, Fz, and Cz. Raw recordings were, first, bandpass filtered (1 Hz to 100 Hz) and then notch filtered (at 50Hz and 100Hz). We re-referenced recordings to the average across channels and used independent component analysis (ICA) to remove any EOG contamination \cite{hyvarinen1999fast}. Recordings were epoched from 0.5 s before to 2 s after the onset of audio feedback, baseline-subtracted (i.e., we subtracted the average of the recording from -0.5 to 0 s from each data point in each epoch), and downsampled to 256 samples per second per channel. We labelled each epoch to indicate the participant’s success or otherwise on the corresponding trial (see \textit{Visual stimuli and feedback}). In total, we recorded three hundred epochs from each participant.

\subsection{Classification and Evaluation}

\subsubsection{Cross-validation} 

To evaluate classifiers, we used 10-fold cross-validation (CV), repeated 10 times (i.e., ``10-by-10-fold” CV). We performed this 10-by-10-fold CV within participant \cite{witten2011data}. 
%On each repeat, the dataset was randomly shuffled into ten parts; each part contained 10\% of the data, and the proportion of the classes (``success” and ``failure”) in each part was approximately equal to the original proportion in the complete dataset. On each of the ten folds in a repeat, the classifier was trained on nine parts of the data (``training" set) and tested on the remaining part of the data (``testing" set). Within each repeat, each part of the data was used only once as the testing set. On each fold, classifier performance was quantified using the testing set,
On each repeat, the dataset was randomly shuffled into ten equally sized parts, each with a balanced proportion of the ``success" and ``failure" classes. For each fold, the classifier was trained on 90\% of the data and its performance was quantified using the remaining 10\%, calculating accuracy in the standard fashion: the proportion of the predicted classes of the epochs that matched the true classes. CV was implemented using Python (version 3.8.3). Specifically, we used scikit-learn (v0.23.1) \cite{scikit-learn}, an open-source library that implements common machine-learning algorithms.

\subsubsection{Within- and between-participant comparison of classifier accuracy} 

To compare classification methods within participant, we used the Nadeau \& Bengio corrected t-test as described by Bouckaert \& Frank \cite{bouckaert2004evaluating}. To compare classification methods across participants, we used a permutation test as follows. For each participant, we computed median accuracy separately for each of the two classifiers being compared, and then calculated the difference between these medians (``classifier 1 minus classifier 2”). This metric was summed across all participants. We then z-scored this metric against a null distribution and computed the p-values. The null distribution comprised 1000 null metrics, each of which was computed in the same way as described above after shuffling the labels (``classifier 1" and ``classifier 2") on accuracies used in the computation.

\subsubsection{Chance-level accuracy} 

%We computed the chance-level accuracy for each participant using a shuffle test based on the benchmark approach 2 (described below). 
We computed the chance-level accuracy for each participant using a shuffle test based on the benchmark approach (described below). 
We randomly shuffled the classes for all epochs, tested the benchmark approach using the CV framework stated above, and recorded the average CV accuracy for 100 random shuffles. Then, we recorded the average and the 97.5 percentile of the shuffled accuracies as the classification chance level and the chance threshold for each participant.

\subsubsection{Riemannian geometry-based feature extraction and classification} 

To extract ErrP features, we adopted methods introduced by Barachant and colleagues \cite{yger2016riemannian,barachant2010riemannian,barachant2014plug}, and used Barachant's Python (pyRiemann, v0.2.6) implementation of these methods \cite{pyriemann}. In brief, on each fold of a 10-fold cross-validation, we used training data to construct covariance matrices. We projected these covariance matrices onto a space tangential to the manifold, defined by the geometric mean of all matrices. This projection vectorized matrices. %\textcolor{red}{The projection vectorizes the matrices, and} %After projection, each matrix was vectorized;
These feature vectors were used for training an L2-regularized logistic regression classifier implemented in scikit-learn \cite{scikit-learn}. To construct covariance matrices, we first windowed our training epochs, using only 100 to 600 ms (i.e., 128 samples per channel), where 0 ms is the onset of feedback. On each channel (32 channels total), we separately averaged ``failure" and ``success" epochs, giving two ``prototype" matrices, each 32 rows-by-128 columns. We concatenated these prototype matrices with each single trial taken from the training set, giving a 96-by-128 ``super trial" matrix. Each super trial was used to compute a 96-by-96 covariance matrix. Parts of this matrix captured the covariance between channels within the given trial; other parts of the matrix captured covariance between the given trial and the prototype trials. %\textcolor{red}{Parts of this covariance matrix captured the covariances between channels within the given trial, and the co-covariances between the given trial and the prototype trials, providing information for classification. }%; covariance matrices were mapped onto the Riemannian tangent space. Mapped covariance matrices were weighted (applying unity weighting to elements on the main diagonal, and 2 to off-diagonal elements) and vectorized (taking only the upper triangle); these feature vectors were used to train the classifier. 
Using these training prototype matrices, we then applied the same procedure to single test trials; we used the feature vectors generated by test trials to assess classifier performance.

\subsubsection{Benchmark feature extraction and classification} 

%To help evaluate the Riemannian geometry-based feature extraction, we developed two benchmarks. For benchmark \#1, for each epoch, we computed a feature vector comprising eight values. Values were (1) mean and (2) standard deviation computed during each of four temporal windows: 100-200 ms, 200-300 ms, 300-400 ms, and 400-600 ms, where 0 ms refers to the onset of audio feedback. We used recordings from only 12 medial-frontal electrodes, corresponding to scalp locations where ErrPs are typically maximal: Fz, Cz, AF3, AF4, F3, F4, FC1, FC2, C3, C4, CP1, CP2. Then we used the same logistic regression classifier as used in the Riemannian geometry-based approach to classify the samples through the features extracted. Benchmark \#2 was identical to \#1, however for benchmark \#2 we used recordings from all 32 electrodes. These benchmarks -- specifically, windowing and electrode subsampling -- were adapted from recent work by others \cite{kakkos2020condition} and \cite{schonleitner2020calibration}.

To help evaluate the Riemannian geometry-based feature extraction, we developed a benchmark. For each epoch, for each electrode, we computed a feature vector comprising eight feature. Features were (1) mean and (2) standard deviation computed during each of four temporal windows: 100-200 ms, 200-300 ms, 300-400 ms, and 400-600 ms, where 0 ms refers to the onset of audio feedback. These features were concatenated and each feature was scaled by its maximum absolute value in the training set. Again, we used the logistic regression classifier to perform classification; this classifier outperformed a range of other classifiers, including a support vector machine, and a linear disciminant analysis (data not shown). %\textcolor{red}{These features were concatenated and each feature was scaled by its maximum absolute value in the training set. Again we used the logistic regression classifier to perform classification - it out-performed a range of other classifiers including support vector machines, linear discriminant analysis, and so on in our tests. }
%Then we used the same logistic regression classifier as used in the Riemannian geometry-based approach to classify the samples using the concatenated feature vectors. 
This benchmark -- specifically, our use of windows -- was adapted from recent work by others \cite{kakkos2020condition} and \cite{schonleitner2020calibration}.

\section{Results}

\subsection{Participants’ behavioural performance}

Participants performed the fixation task capably. To estimate each participant's threshold, we averaged across blocks the last 10 reversals of each block’s staircase, and then transformed the contrasts back into non-logged Weber contrasts \cite{cobo1994selectivity}. Thresholds and performance are shown in TABLE \ref{tab:behaviour}.

\begin{table}[htpb]
% \setlength\extrarowheight{1.5pt}
% \captionsetup{margin=1em}
\caption{The discrimination thresholds (in Weber contrast \%) and behavioural performance (\% correct behavioural responses) for all participants (P1 to P7). The percentage of correct responses was calculated across all five blocks.
}
\label{tab:behaviour}
\centering
% \begin{center}
\resizebox{0.48\textwidth}{!}{
\begin{tabular}{lrrrrrrr}
\toprule
                         & P1    & P2    & P3    & P4    & P5    & P6    & P7    \\
\midrule
Discrimination threshold & 0.338 & 0.779 & 0.542 & 1.129 & 0.161 & 0.269 & 0.382 \\
Behavioural performance  & 57.0  & 55.7  & 57.3  & 55.7  & 59.7  & 59.7  & 57.3  \\
\bottomrule
\end{tabular}
}
% \end{center}
\end{table}

\subsection{ERP components}

\begin{figure}[thpb]
      \centering
      \centerline{\includegraphics[width=0.45\textwidth]{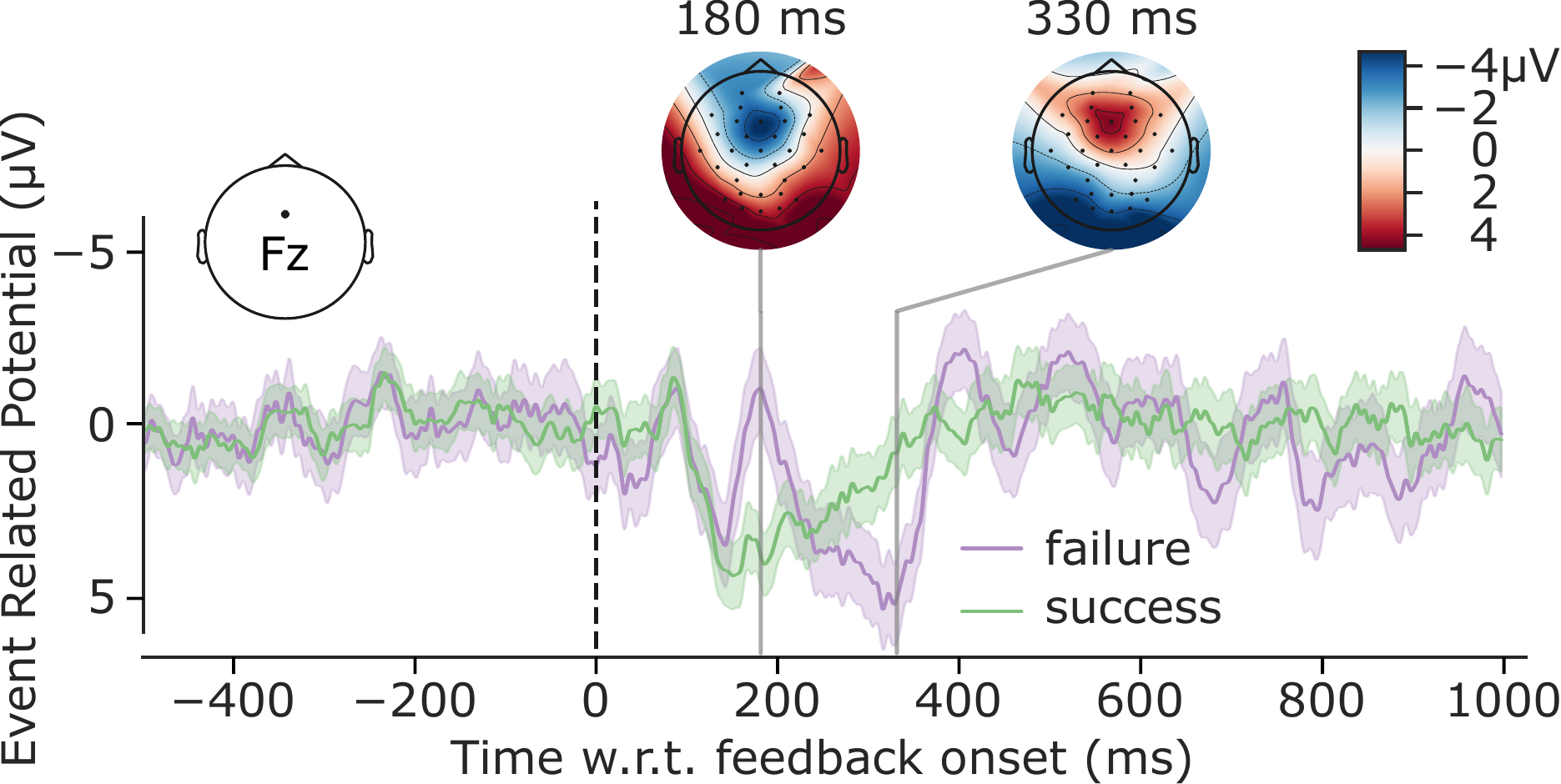}}
      \caption{Example ERPs, participant P5. ERPs recorded on electrode Fz (inset), showing the average of trials on which the participant received feedback indicating success on the discrimination task (green) and feedback indicating failure (purple). Shaded regions show 95\% confidence intervals computed via bootstrapping. The scalp maps show ``failure minus success” at 180 ms (left) and 330 ms (right). For this participant, the negativity at 180 ms was localised to a frontal-central region surrounding electrode Fz; the positivity at 330 ms showed similar spatial organisation.
}
        \label{fig:example_erps}
  \end{figure}

We recorded robust ERPs from all participants. In all participants we observed a FRN at the fronto-central scalp areas (centered at Fz) in failure epochs compared with success epochs (failure minus success), peaking between 100 to 200 ms. Following this negativity, in 5 of 7 participants, we also observed a frontally distributed positivity which peaked around 300 ms (P3A \cite{ullsperger2014neural,polich2007updating}). These two ERP components are exemplified in Fig. \ref{fig:example_erps}. These observations were broadly consistent with the ErrP described in the literature \cite{chavarriaga2014errare,ullsperger2014neural}, which we discuss below (see \textit{Discussion}).

\addtolength{\textheight}{-0.5cm}   % This command serves to balance the column lengths
%                                   % on the last page of the document manually. It shortens
%                                   % the textheight of the last page by a suitable amount.
%                                   % This command does not take effect until the next page
%                                   % so it should come on the page before the last. Make
%                                   % sure that you do not shorten the textheight too much.
% \begin{figure}[thpb]

% % \centering
%     (A)
%     \vspace{2mm}
    
%     \begin{subfigure}{.45\textwidth}
%     \centerline{
%     \includegraphics[width=\textwidth]{figures/paper_figures_beta_ERP_Fz.pdf}}
%     \end{subfigure}
    
%     \vspace{4mm}
    
%     (B)
%     \vspace{-3mm}
    
%     \begin{subfigure}{.45\textwidth}
%     \centerline{
%     \includegraphics[width=\textwidth]{figures/paper_figures_beta_ERP_all_chs.pdf}}
%     \end{subfigure}

% \caption{\textbf{Example ERPs, participant P5.} A) ERPs recorded on electrode Fz (inset), showing the average of trials on which the participant received feedback indicating success on the discrimination task (green) and feedback indicating failure (purple). Shaded regions show 95\% confidence intervals computed via bootstrapping. Feedback was provided at time = 0 ms. B) Average ERPs for all 32 channels. The scalp maps show ``failure minus success” at 180 ms (left) and 330 ms (right). For this participant, the negativity at 180 ms was localised to a medial-frontal region surrounding electrode Fz; the positivity at 330 ms showed similar spatial organisation.
% }

% \label{fig:example_erps}

% \end{figure}

\subsection{Classification approach performance}

\begin{figure}[thpb]
      \centering
      \centerline{\includegraphics[width=0.45\textwidth]{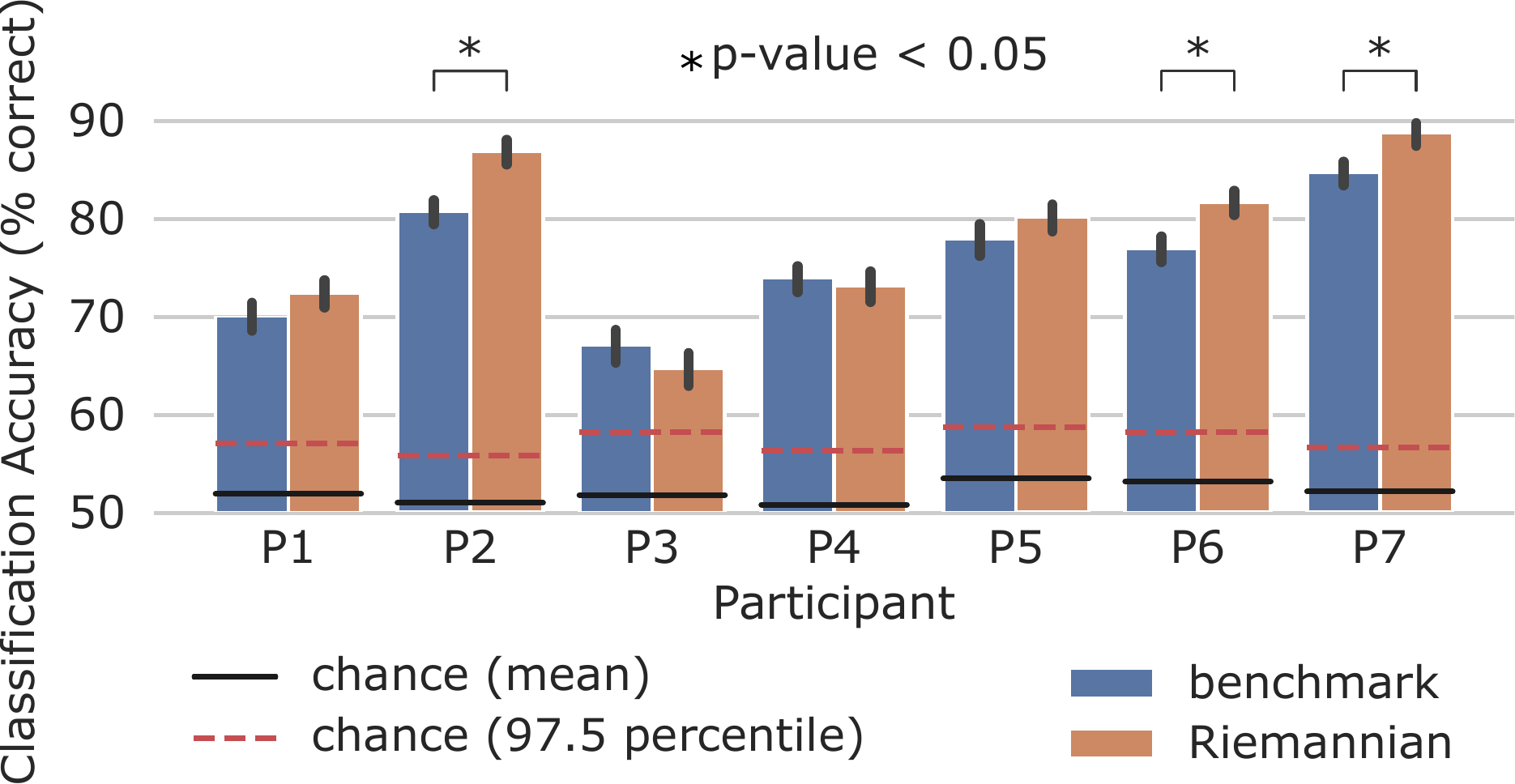}}
      \caption{Comparison of cross-validated classifier accuracy. In 3 of 7 experimental participants, the performance of the Riemannian geometry-based method was statistically significantly greater than that of the benchmark.
}
        \label{fig:score_comparison}
  \end{figure}

We used 10-by-10-fold cross-validation to quantify the accuracy of all classification approaches (see \textit{Cross-validation}); all approaches performed at rates above chance. 
%In none of the participants did we see a statistically significant difference between the performance of benchmark 1 (medial-frontal electrodes) and 2 (all electrodes); however, across all participants, the accuracy averaged across participants for benchmark 2 = 75.9\% versus benchmark 1 = 74.4\%, and the permutation test suggests a significant difference between the two approaches (z-statistics=2.745, p=0.006). Data are shown in Fig. \ref{fig:score_comparison}.
%
%Overall, the Riemannian geometry-based approach outperformed both benchmark approaches. For experimental participants P2, P6, and P7, we saw a statistically significant difference between the performance of the Riemannian approach and both benchmarks: P2, 86.8\% versus 79.0\% (benchmark 1; t-statistics=2.701, p=0.008) and 80.7\% (benchmark 2; t-statistics=2.326, p=0.022); P6, 81.6\% versus 76.6\% (benchmark 1; t-statistics=1.985, p=0.04996) and 76.9\% (benchmark 2; t-statistics=2.136, p=0.035); P7, 88.7\% versus 84.7\% (benchmark 1; t-statistics=2.439, p=0.017) and 83.8\% (benchmark 2; t-statistics=2.016, p=0.047). In two other participants (P1 and P5), the Riemannian approach outperformed both benchmarks, but this difference did not reach statistical significance. Across all participants, the Riemannian approach also reached an overall accuracy of 78.2\%, showing a significantly better overall performance: versus 74.4\% by benchmark 1 (z-statistics=5.920, p=3.212e-09); and versus 75.9\% by benchmark 2 (z-statistics=4.028, p=5.632e-05). Data are shown in Fig. \ref{fig:score_comparison}.
Overall, the Riemannian geometry-based approach outperformed the benchmark. For experimental participants P2, P6, and P7, we saw significantly higher performance using the Riemannian approach than that of the benchmark: P2, 86.8\% versus 80.7\% (t=2.326, p=0.022); P6, 81.6\% versus 76.9\% (t=2.136, p=0.035); P7, 88.7\% versus 83.8\% (t=2.016, p=0.047). In two other participants (P1 and P5), the Riemannian approach outperformed the benchmark, but the differences did not reach statistical significance. Across all participants, the Riemannian approach also reached an overall accuracy of 78.2\% which was statistically significantly greater than the benchmark’s performance of 75.9\% by the benchmark (z=4.028, p=5.632e-05). Data are shown in Fig. \ref{fig:score_comparison}.

%   \begin{figure}[thpb]
%       \centering
%       \centerline{\includegraphics[width=0.45\textwidth]{figures/paper_figures_beta_score_comparison.pdf}}
%       \caption{\textbf{Comparison of cross-validated classifier accuracy.} In 3 of 7 experimental participants, the performance of the Riemannian geometry-based method was statistically significantly greater than that of both benchmark classifiers.
% }
%         \label{fig:score_comparison}
%   \end{figure}

%\addtolength{\textheight}{-0.3cm}   % This command serves to balance the column lengths

\section{Discussion} \label{sec:discussion}

Our results demonstrate that the Riemannian geometry-based approach to classifying the FRN outperforms a traditional approach. Because both of these approaches used the same classifier (logistic regression), it seems that the Riemannian approach was better able to extract the salient features contained in the FRN. A further advantage of the Riemannian approach is that it represents both spatial and temporal information without much feature engineering (i.e., the postulation of features and specification of temporal windows, as is necessary in the traditional approach).

ErrPs elicited by error feedback have a characteristic morphology; a frontocentral FRN appears between 200 and 300 ms after feedback, followed by a P3A appearing after 300 ms \cite{ullsperger2014neural}. In our recordings, we observed a positivity in five of seven participants (P2, P4 through P7) after 300 ms. We observed a negativity surrounding site Fz (e.g., Fig. \ref{fig:example_erps}) in all participants. However, this peaked somewhat early, between 100 and 200 ms, that is, where one might expect to find the N1 and P2 components of the auditory evoked response (AER) \cite{naatanen1987n1}. Because we provided  feedback using pure tones at 700 and 200 Hz, it is possible that frequency-related differences in N1/P2 contributed to the negativity we observed.  However we suspect this contribution is small. Picton et al. \cite{picton1978human} measured N1 and P2 amplitudes as a function of frequency, using tone bursts of 250, 500, and 1000 Hz at 87 ± 3 dB SPL. Picton’s data indicate that our use of 200 and 700 Hz for feedback may contribute a small negativity to our recordings at around 100 to 175 ms. However, the tone bursts used by Picton were high-intensity compared to ours, and may not accurately model AERs in our participants. In ongoing work, we aim to separate the FRN from any AERs.

\bibliographystyle{IEEEtran}
\bibliography{citations}

\end{document}